\documentclass[conference]{IEEEtran}
\IEEEoverridecommandlockouts
\usepackage{hyperref}
\usepackage{cite}
\usepackage{amsmath,amssymb,amsfonts,pifont}
\usepackage{algorithmic}
\usepackage{multirow,graphicx,colortbl}
\usepackage{textcomp}
\usepackage{xcolor}
\definecolor{lightgray}{gray}{0.9}
\def\BibTeX{{\rm B\kern-.05em{\sc i\kern-.025em b}\kern-.08em
    T\kern-.1667em\lower.7ex\hbox{E}\kern-.125emX}}
\begin{document}

\title{Decoupling Feature Representations of Ego and Other Modalities for Incomplete Multi-modal Brain Tumor Segmentation\\
\thanks{$^\dag$: Co-first author, $^*$: Corresponding authors.}
}

\author{\IEEEauthorblockN{Kaixiang Yang$^{a,b,\dag}$, Wenqi Shan$^{a,\dag}$, Xudong Li$^c$, Xuan Wang$^c$, Xikai Yang$^d$, Xi Wang$^d$,\\
Pheng-Ann Heng$^d$, Qiang Li$^{a,*}$, Zhiwei Wang$^{a,*}$}
\IEEEauthorblockA{\textit{$^a$Wuhan National Laboratory for Optoelectronics, Huazhong University of Science and Technology} \\
\textit{$^b$College of Life Science and Technology, Huazhong University of Science and Technology} \\
\textit{$^c$Union Hospital, Tongji Medical College, Huazhong University of Science and Technology}\\
\textit{$^d$The Chinese Univsersity of Hong Kong}\\
\{liqiang8, zwwang\}@hust.edu.cn}
}

\maketitle

\begin{abstract}
Multi-modal brain tumor segmentation typically involves four magnetic resonance imaging (MRI) modalities, while incomplete modalities significantly degrade performance. Existing solutions employ explicit or implicit modality adaptation, aligning features across modalities or learning a fused feature robust to modality incompleteness. They share a common goal of encouraging each modality to express both itself and the others. However, the two expression abilities are entangled as a whole in a seamless feature space, resulting in prohibitive learning burdens. In this paper, we propose DeMoSeg to enhance the modality adaptation by \textbf{De}coupling the task of representing the ego and other \textbf{Mo}dalities for robust incomplete multi-modal \textbf{Seg}mentation. The decoupling is super lightweight by simply using two convolutions to map each modality onto four feature sub-spaces. The first sub-space expresses itself (Self-feature), while the remaining sub-spaces substitute for other modalities (Mutual-features). The Self- and Mutual-features interactively guide each other through a carefully-designed Channel-wised Sparse Self-Attention (CSSA). After that, a Radiologist-mimic Cross-modality expression Relationships (RCR) is introduced to have available modalities provide Self-feature and also `lend' their Mutual-features to compensate for the absent ones by exploiting the clinical prior knowledge. 
The benchmark results on BraTS2020, BraTS2018 and BraTS2015 verify the DeMoSeg's superiority thanks to the alleviated modality adaptation difficulty. Concretely, for BraTS2020, DeMoSeg increases Dice by at least 0.92\%, 2.95\% and 4.95\% on whole tumor, tumor core and enhanced tumor regions, respectively, compared to other state-of-the-arts. Codes are at {\href{https://github.com/kk42yy/DeMoSeg}{https://github.com/kk42yy/DeMoSeg}}
\end{abstract}


\section{Introduction}
Glioma is a common malignant tumor in the central nervous system with high mortality. For imaging diagnosis of glioma, T1-weighted (T1), contrast enhanced T1-weighted (T1ce), T2-weighted (T2) and Fluid Attenuation Inversion Recovery (FLAIR) are common magnetic resonance imaging (MRI) modalities.
Specifically, T1 and T1ce are usually for assessing the tumor core, while T2 and FLAIR are used for edema~\cite{CKD-TMI}. To help accurate and effective diagnosis, numerous deep learning-based methods for brain tumor segmentation~\cite{eoformer,CKD-TMI,EdgeInFusion} have been proposed, and they typically rely on all above modalities.
However, image corruption, scanning protocols, conditions of patients and other factors~\cite{GSS_union2,MAVP_fusion7,MRI-missingreason1} often cause incomplete modality collection, degrading the performance of the above methods significantly. Therefore, algorithms for incomplete multi-modal segmentation are of great demand in clinical practice.

Incomplete multi-modal brain tumor segmentation aims at making use of available modalities to segment brain tumor, achieving the results similar to, or as good as using the full modalities. 
The existing efforts mainly adopt the paradigm of modality adaptation, in either explicit~\cite{Introduction_Generate,seperate1,ACN_seperate2,seperate3,CrossKD_union1} or implicit~\cite{HeMIS_fusion1,U-HVED_fusion2,RSeg_fusion3,RFNet_fusion4,UNet-MFI_fusion5,mmFormer_fusion6,MAVP_fusion7,GSS_union2} manners. 
The explicit modality adaptation often tries to restore the missing modalities using a modality-to-modality translation~\cite{Introduction_Generate}, or mimics the full-modality features using knowledge transfer~\cite{seperate1,ACN_seperate2,seperate3,CrossKD_union1}.
For instance, Wang et al.~\cite{CrossKD_union1} proposed cross-modal knowledge distillation to handle missing modality, where features of missing modalities are generated by averaging the available modalities' features.

The implicit modality adaptation fuses the features from different modalities and utilizes a shared decoder to give the final segmentation prediction based on the fused feature. It is trained using a modality perturbation strategy, \textit{i.e.}, randomly dropping some modalities by replacing the corresponding features with zeros, to encourage the shared decoder to be invariant to the fused feature of incomplete modalities. An auxiliary regularizer is often employed to facilitate extracting features from the available modalities as informative as those from the full modalities. 
For instance, Ding et al.~\cite{RFNet_fusion4} proposed a region-aware fusion module to aggregate features from each modality encoder branch for incomplete scenarios. 
Qiu et al.~\cite{GSS_union2} proposed a group self-support framework for inter-modal information complementarity, which served as an extra regularizer for better feature fusion during training. 

Although the strategy of modality adaptation differs, current methods for incomplete multi-modal segmentation share a common goal, that is, they all expect the feature from each modality to contain information of both ego-modality and other-modalities, concurrently. These are actually two different capabilities, \textit{i.e.}, self-expression and mutual-expression, but existing methods mostly entangle them in the same feature space and assign the two kinds of information onto each feature dimensionality, equally.

In this paper, we argue that such information entanglement of ego-modality and other-modalities creates a heavy burden for feature learning, and reduces the effectiveness of modality adaptation. In response, we aim to separate the features into Self-feature and Mutual-features parts for each modality, and instantiate this motivation as a new framework, named \textbf{De}coupled \textbf{Mo}dality features based incomplete multi-modal \textbf{Seg}mentation (DeMoSeg). Unlike the previous methods adopting a giant encoding branch to learn a highly-entangled feature for each modality, DeMoSeg enjoys the alleviated task difficulty and thus simply uses two convolutional layers to map each modality onto four feature sub-spaces. These sub-spaces have different responsibilities: the first is Self-feature for expressing the original modality itself, and each of the others is Mutual-features for simulating one of the other modalities. Also, a new layer named Channel-wised Sparse Self-Attention (CSSA) is attached to permit a mutual guidance across the sub-spaces for a more effective feature decoupling.

On top of the decoupling, DeMoSeg introduces a feature compensation step, which fully exploits clinical prior knowledge to construct a pseudo complete multi-modal feature by use of available modalities. The decoupled feature sub-spaces permit to mimic a radiologist's interpretation on incomplete multi-modal MRI data. For example, the T2-expressed FLAIR-specific Mutual-features will take over if FLAIR is absent, while T1 and T1ce can provide the mutual-expression feature for each other if one of them is absent. 

In summary, our main contributions are listed as follows:
\begin{itemize}
    \item We propose a novel DeMoSeg for incomplete multi-modal brain tumor segmentation, which efficiently decouples the Self-feature and Mutual-features of each modality to reduce the learning burden in modality adaptation. Thus, DeMoSeg selects Mutual-features from better suited available modalities to fill the void ones under different modality perturbations, in which the radiologist's prior knowledge can be easily implanted.
    \item We propose a new attention layer called Channel-wised Sparse Self-Attention (CSSA) to enable cross-guidance between the Self- and Mutual-features, but meanwhile prevent them from being coupled again. Moreover, permitted by the decoupled features, we propose a Radiologist-mimic Cross-modality expression Relationships (RCR) to construct a pseudo full-modality feature effectively.
    \item We conduct extensive experiments on the benchmark datasets BraTS2020, BraTS2018 and BraTS2015. Thanks to efficient feature decoupling and knowledge driven feature compensation, DeMoSeg outperforms the state-of-the-art methods in Dice by at least 0.92\%, 2.95\% and 4.95\% on whole tumor, tumor core and enhanced tumor for BraTS2020, respectively.
\end{itemize}

\section{Methodology}
\begin{figure*}
\centering
\includegraphics[width=\textwidth]{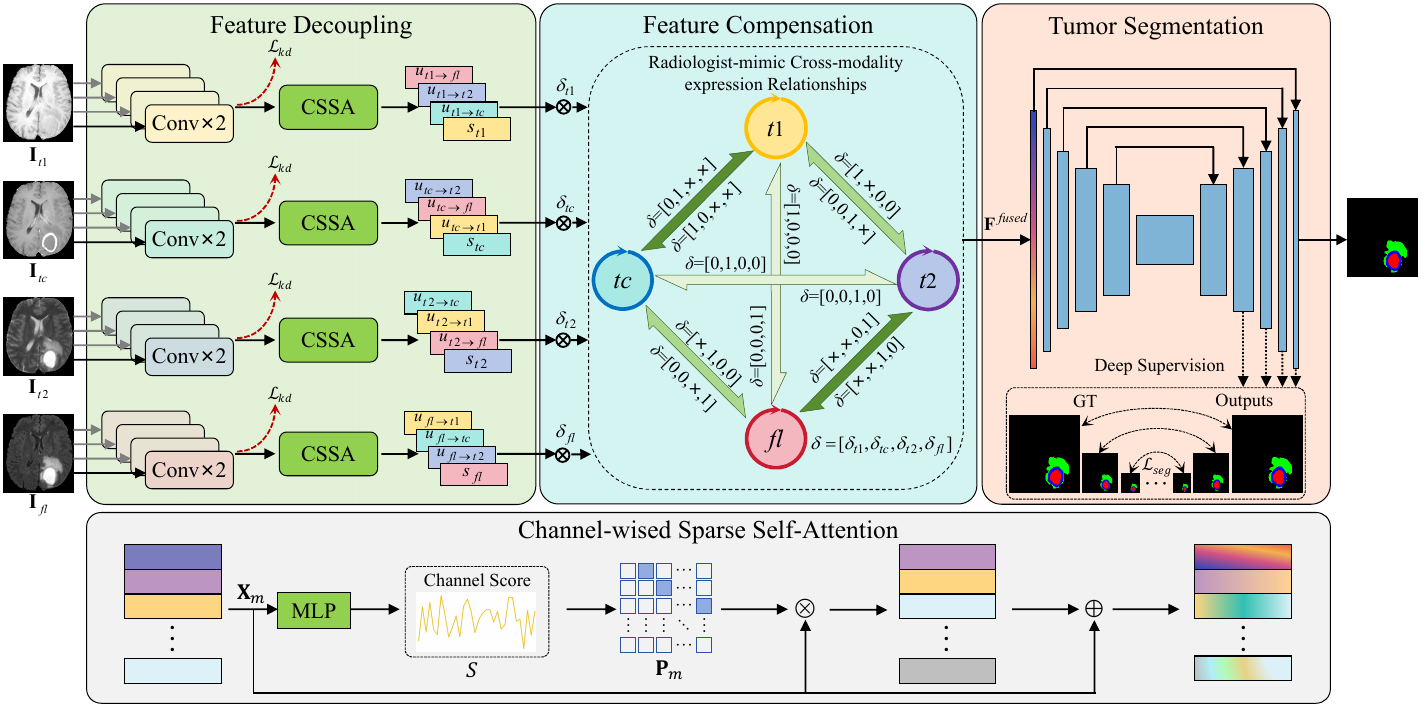}
\caption{The overall framework of DeMoSeg, which consists of three main parts, (1) Feature Decoupling, (2) Feature Compensation and (3) Tumor Segmentation. $s_m$ represents the Self-feature of modality $m$, and $u_{m \rightarrow l} $ means Mutual-features of modality $m$ representing modality $l$. The alignment constraint $\mathcal{L}_{kd}$ and segmentation network loss $\mathcal{L}_{seg}$ are formulated as Eq.~\ref{klloss} and Eq.~\ref{segloss}, respectively. In Feature Compensation, the bidirectional arrows' color intensity corresponds to the priority level, with darker colors indicating higher priority. Modality indicator $\delta_m \in [0,1]$ represents the modality $m$ is missing or not, for instance, $\delta=[0,0,1,\times]$ means $t1$ and $tc$ are missing, $t2$ is available, $fl$ can be both.}
\label{fig1_overall_stracture}
\end{figure*}

Fig.~\ref{fig1_overall_stracture} shows the framework of our proposed DeMoSeg. It consists of three major components, \textit{i.e.}, (i) feature decoupling of self and mutual expression for each modality, (ii) feature compensation based on clinical prior knowledge, and (iii) a U-Net backbone for tumor segmentation. In the following, we explain each component.

\subsection{Feature Decoupling of Self and Mutual Expression}
Current methods mostly encode each modality into highly entangled self and mutual expression parts, requiring giant encoders and an auxiliary regularizer.

\subsubsection{Feature Partition}
We propose feature decoupling to separate the modality features into two parts, Self-feature and Mutual-features, which is easier than learning entangled features, so a few parameters is enough, and more effort can be devoted into the subsequent tumor segmentation network. We denote the input modality $m$ with the shape of $D\times H\times W$ as $\mathbf{I}_m\in \mathbb{R}^{1 \times D \times H \times W}$, where $ m \in \mathcal{M}= \{t1, tc, t2, fl\}$, $t1, tc, t2$ and $fl$ represent T1, T1ce, T2 and FLAIR, respectively. Our feature decoupling is to map the input $\mathbf{I}_m$ onto four sub-spaces, 
\textit{i.e.}, $\mathbf{I}_m \mapsto \{s_m, u_{m\rightarrow(\mathcal{M}\backslash\{m\})} \}$, where the \textbf{Self-feature} $ s_m $ presents intrinsic features of modality $m$ and the \textbf{Mutual-features} $u_{m\rightarrow(\mathcal{M}\backslash\{m\})}$ are ready to compensate features for missing modalities.

We intentionally assign responsibilities of expressing different modalities to $ s_m $ and $u_{m\rightarrow(\mathcal{M}\backslash\{m\})}$ as depicted in Fig.~\ref{fig1_overall_stracture}. Specifically, we firstly employ two successive $3\times 3\times 3$ convolution layers to obtain intermediate-decoupled features, $s'_m \in \mathbb{R}^{C \times D \times H \times W}$ and $u'_{m\rightarrow(\mathcal{M}\backslash\{m\})} \in \mathbb{R}^{3C \times D \times H \times W}$, and $C$ is set to 8 according to the channel number of our tumor segmentation backbone. Intuitively, if the extracted $u'_{(\mathcal{M}\backslash\{m\})\rightarrow m}$ are consistent with the $s'_m$, they could perform better in the absence of certain modalities. Hence, we utilize knowledge distillation~\cite{KnowledgeDistillation} as a constraint to align features from $u'_{(\mathcal{M}\backslash\{m\})\rightarrow m}$ to $s'_m$, which is detailed in \hyperlink{ii-c}{Sec II.C}.

Next, we introduce a Channel-wised Sparse Self-Attention (CSSA) layer to permit the Self- and Mutual-features to lightly interact with each other for better guiding the feature learning while avoiding features re-entanglement. 

\subsubsection{Channel-wised Sparse Self-Attention}
Traditional self-attention creates dense attention map, aggregating the Self- and Mutual-features too much and making them recoupled again. Therefore, we aim to generate a sparse attention matrix to avoid recoupling. Our core idea is to regress a score for each channel, and sort the channels’ scores to build an attention matrix. According to the scores, we intentionally associate the first channel with that of the highest score, the second channel with that of the second highest score, and so on. That is, each channel is only associated with a unique target channel and vice versa, which means that a bijective mapping between original and sorted channels can be built to be equivalent to a sparse and full-rank attention matrix.

In detail, we concatenate the intermediate features $s'_m$ and $u'_{m\rightarrow(\mathcal{M}\backslash\{m\})}$ from the preliminary decoupling as $ \mathbf{X}_m \in \mathbb{R}^{C_1 \times D \times H \times W} $, and utilize a global average pooling (GAP) and two MLP layers to obtain scores $S \in \mathbb{R}^{C_1}$ for every channel, where $C_1=4C$ and the score $S$ can be formulated as:
\begin{equation}
    S = \mathbf{MLP}(\textbf{GAP}(\mathbf{X}_m))
\end{equation}

Then we sort $S$ and have an indicator $Q$, where $Q(i)=j$, means that the $j$-th channel gets the $i$-th highest score, and $0 \le i,j < C_1$ ($i=0$ means the highest). Next, we convert $Q(i)$ into a one-hot vector $\mathbf{e}_{Q(i)}=\mathbf{e}_{j}$, and construct a permutation matrix $\mathbf{P}_m$:
\begin{equation}
    \mathbf{P}_m=
    \begin{pmatrix}
        \mathbf{e}_{Q(0)}^{\rm{T}};& \mathbf{e}_{Q(1)}^{\rm{T}}; & \dots ; & \mathbf{e}_{Q(C_1-1)}^{\rm{T}}
    \end{pmatrix} ^{\rm{T}} \in \mathbb{R}^{C_1 \times C_1}
\end{equation}

Therefore, the final CSSA result can be written as:
\begin{equation}
    CSSA(\mathbf{X}_m)= \mathbf{X}_m + \mathbf{P}_m * \mathbf{X}_m
\end{equation}
where $\mathbf{P}_m$ is sparse to avoid entanglement again among channels, and can act as an attention matrix. After CSSA, we evenly re-separate the output into $\{s_m, u_{m\rightarrow(\mathcal{M}\backslash\{m\})}\}$ for the following feature compensation.

\subsection{Feature Compensation based on Clinical Knowledge}
Incomplete modality can be formulated as subsets of $\mathcal{M} = \{t1, tc, t2, fl\}$ except $\varnothing$. To handle different scenarios, a random perturbation is employed using an indicator, $\delta = [\delta_{t1}, \delta_{tc}, \delta_{t2}, \delta_{fl}]$, $\delta_m \in [0, 1]$. If $\delta_{m} = 0$, instead of using zeros to fill the void, we carefully design a Radiologist-mimic Cross-modality expression Relationships (RCR) to generate pseudo full-modality features $\triangleq \mathbf{F}^{fused}$.

\begin{figure*}[htbp]
\centering
\includegraphics[width=\textwidth]{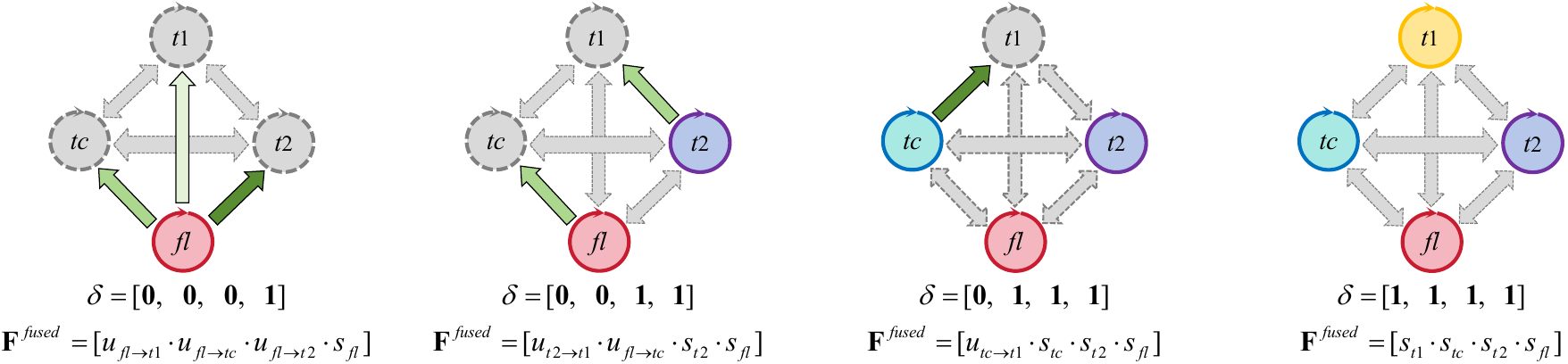}
\caption{Examples of radiologist-mimic cross-modality feature compensation strategy.}
\label{fig2_compensation_examples}
\end{figure*}

Clinically, radiologists usually combine $t1$ and $tc$ to diagnose tumor core, while $t2$ and $fl$ are combined to diagnose edema~\cite{CKD-TMI}, which are the \textit{primary} relationships. Besides, $tc$ and $fl$ perform better in single-modality segmentation~\cite{CrossKD_union1} and $t1$ and $t2$ show similar physical property, which means that these two pairs can compensate each other and have the \textit{secondary} relationships if the primary is invalid.
Based on the above analysis, the RCR is created as visualized in Fig.~\ref{fig1_overall_stracture}. For full-modality, we obtain $\mathbf{F}^{fused}=[s_{t1} \cdot s_{tc}\cdot s_{t2}\cdot s_{fl}]$, where $[\cdot]$ means feature concatenation along channel dimension. When some modalities are missing, \textit{i.e.}, $\delta$ is not all 1, we can utilize RCR to construct pseudo full-modality features. Fig.~\ref{fig2_compensation_examples} shows some specific construction examples, for instance, the second example means that when $t1$ and $tc$ are missing, $t2$ and $fl$ will be used to compensate $t1$ and $tc$, respectively, resulting in $\mathbf{F}^{fused}=[u_{t2\rightarrow t1}\cdot u_{fl\rightarrow tc}\cdot s_{t2}\cdot s_{fl}]$.

Moreover, the feature decoupling and compensation can also be implicitly supervised through segmentation results during training. The full modality feature $\mathbf{F}^{fused}$ is obtained by concatenating the 4 modalities’ $s_m$ along channel by a specific order, \textit{i.e.}, $[s_{t1} \cdot s_{tc}\cdot s_{t2}\cdot s_{fl}]$. If some are missing, their RCR-corresponding modalities’ Mutual-features will take over their places (\textit{i.e.}, channels) in the full modality feature. By doing so, the substitute features (\textit{i.e.}, Mutual) will be forced to pretend to be the original features (\textit{i.e.}, Self) in their occupied channels. Thus, the different expression abilities of Mutual-features can be decoupled and learned well.

\subsection{Tumor Segmentation and Overall Training}\hypertarget{ii-c}{}
As shown in the rightmost part of Fig.~\ref{fig1_overall_stracture}, to segment brain tumor using $\mathbf{F}^{fused}$, we adopt 3D U-Net~\cite{U-Net,3DUNet,nnunet} as backbone, and replace the first encoding layer with our proposed feature decoupling and compensation steps. 
Thanks to our light-weight enabling technique, we can employ a more powerful UNet backbone with 6 feature scales, 5 times down-sampling and a maximum of 320 channels. In the remaining encoder and decoder, we use convolution with stride 2 for down-sampling features and transpose convolution for up-sampling features, respectively. Besides, the second encoding layer of encoder takes an input with 32 channels, and we have 4 modalities, $C$ in feature partition is set to 8.

In feature decoupling, knowledge distillation has been utilized to align features:
\begin{equation}\label{klloss}
    \mathcal{L}_{kd} = \sum_{m\in\mathcal{M}} \sum_{n\in\mathcal{M}\backslash\{m\}} \mathbf{KL}(\sigma(\frac{u'_{n \rightarrow m}}{t}) || \sigma(\frac{s'_m}{t}))
\end{equation}
where $\sigma$ represents the softmax operation, $t$ is temperature in knowledge distillation and set to $1$, and $\mathbf{KL}$ denotes the Kullback-Leibler divergence. 
For tumor segmentation network training, we use Dice Loss and Cross-Entropy Loss as loss function:
\begin{equation}\label{segloss}
        \mathcal{L}_{seg}=\frac{1}{K}\sum_{k=0}^{K-1}(1-\frac{2\sum_n g_n^kp_n^k}{\sum_n g_n^k+\sum_n p_n^k})-\frac{1}{N}\sum_{k=0}^{K-1}\sum_{n=0}^{N-1}g_n^k\log p_n^k
\end{equation}
where $K$, $N$ represent the number of classes, voxels, respectively, and $g_n^k$ and $p_n^k$ are the $n$-th ground-truth and prediction values for the $k$-th class. We also apply the deep supervision strategy~\cite{DeepSupervision}, consistent with the comparison methods, to improve model performance. 
The overall loss $\mathcal{L}$ can be formulated as:
\begin{equation}
    \mathcal{L}=\mathcal{L}_{seg} + \mathcal{L}_{kd}
\end{equation}

\section{Experiments}

\subsection{Materials and Details}
\textbf{Datasets, comparison methods and evaluation metric.}
Our experiments are mainly conducted on Multi-modal Brain Tumor Segmentation Challenge (BraTS) Datasets~\cite{BraTS}, \textit{i.e.}, BraTS2020, BraTS2018 and BraTS2015, which contain all the four MRI modalities.   
All cases have been skull-stripped, co-registered and interpolated to the same resolution by organizers. Different brain anatomic structures are combined into three sub-regions, Whole Tumor (WT), Tumor Core (TC) and Enhancing Tumor (ET). 
We compare our method with other state-of-the-art methods, RFNet~\cite{RFNet_fusion4}, mmFormer (mmF)~\cite{mmFormer_fusion6}, MAVP~\cite{MAVP_fusion7} and GSS~\cite{GSS_union2}. 
We follow the data split completely consistent with the above comparison methods, \textit{i.e.}, 219, 50, 100 cases on BraTS2020, 199, 29, 57 cases on BraTS2018, and 242, 12, 20 cases on BraTS2015, for training, validation and test, respectively. We employ Dice Similarity Coefficient (DSC) as the evaluation metric.

\textbf{Implementation Details.}
To train DeMoSeg, we firstly crop the black background and normalize each modality to zero mean and unit variance in brain region. Patches of $128 \times 128 \times 128$ are randomly cropped. Sliding window strategy is used for inference. Data augmentation includes random rotations, blurring, noise and mirror flipping. The whole framework is trained for 1000 epochs using a batch size of 2 on a single NVIDIA GeForce RTX 4090 24GB GPU. Stochastic gradient descent (SGD) optimizer with an initial learning rate of 0.01 and a poly learning rate policy is used for optimization. A post-processing strategy is used to mitigate false alarms of ET following~\cite{U-HVED_fusion2,RFNet_fusion4,MAVP_fusion7,GSS_union2}, where ET less than 500 voxels is reclassified as non-enhancing tumor. 

\subsection{Comparison with the state-of-the-arts}
\textbf{Quantitative Comparison.} The segmentation results of 15 distinct missing modality scenarios on BraTS2020, BraTS2018 and BraTS2015 are listed in Table~\ref{tab3_result_comparison2020},~\ref{tab3_result_comparison2018} and~\ref{tab3_result_comparison2015}. We quote the reported results of RFNet from~\cite{RFNet_fusion4}, mmF and MAVP from~\cite{MAVP_fusion7}, GSS from~\cite{GSS_union2} for fair comparison. From Table~\ref{tab3_result_comparison2020}, our proposed DeMoSeg obtains the highest average DSC on all three sub-regions, exceeding the second-best method, \textit{i.e.}, GSS, in the average DSC by relative 0.92\%, 2.95\% and 4.95\% on WT, TC and ET, respectively. 
Scenarios with more than one missing modalities are difficult to handle because of the scarce and limited information provided by available modalities. Results from the first ten rows of Table~\ref{tab3_result_comparison2020} show the aforementioned exceedingly arduous scenarios, and our DeMoSeg achieves a relatively higher DSC than GSS, \textit{i.e.}, 1.18\%, 3.40\% and 6.38\% for WT, TC and ET, respectively. This demonstrates that DeMoSeg's feature decoupling and compensation based on clinical knowledge can enhance the ability of extracting features from available modalities and thereby maximize the utilization of them. We can draw similar conclusions from Tables~\ref{tab3_result_comparison2018} and~\ref{tab3_result_comparison2015}, which show that DeMoSeg achieves significant improvements compared to previous SOTA methods. We also notice that our method is not the best for ET on BraTS2015. The reason could be that the ET regions are sometimes too small or even nonexistent, and the proportion of these cases is relatively high on BraTS2015 due to the overall smallest-scale testset. For cases with small ET, our prediction might be removed by post-processing, and lead to abnormal DSC situations, \textit{i.e.}, $\textrm{DSC}=0$.

\begin{table*}
\caption{DSC(\%) comparison with RFNet~\cite{RFNet_fusion4}, mmFormer (mmF)~\cite{mmFormer_fusion6}, MAVP~\cite{MAVP_fusion7} and GSS~\cite{GSS_union2} on BraTS2020. Available modalities are denoted by $\bullet$, and missing modalities are represented by $\circ$. The best DSC results are indicated in bold, and the second-best DSC results are underlined.}
\label{tab3_result_comparison2020}
\centering
\resizebox{\linewidth}{!}{
\begin{tabular}{cccc|ccccc|ccccc|ccccc}
\hline
\multicolumn{4}{c|}{Modality}  & \multicolumn{5}{c|}{WT}  & \multicolumn{5}{c|}{TC}  & \multicolumn{5}{c}{ET}  \\ \hline
$fl$        & $t1$        & $tc$        & $t2$        & RFNet & mmF   & MAVP  & GSS   & Ours           & RFNet & mmF   & MAVP  & GSS   & Ours           & RFNet & mmF   & MAVP           & GSS            & Ours           \\ \hline
$\circ$        & $\circ$         & $\circ$          & $\bullet$         & 86.05 & 86.14 & 86.67 & \underline{87.62} & \textbf{88.31} & 71.02 & 70.86 & 70.97 & \underline{72.26} & \textbf{73.83} & 46.29 & 46.34 & 47.22          & \underline{51.28}          & \textbf{53.84} \\
$\circ$        & $\circ$         & $\bullet$          & $\circ$         & 76.77 & 78.47 & 79.46 & \underline{80.14} & \textbf{82.10} & 81.51 & 83.99 & \underline{84.26} & 83.38 & \textbf{85.70} & 74.85 & 80.08 & \underline{81.36} & 78.62  & \textbf{81.42}          \\
$\circ$        & $\bullet$         & $\circ$          & $\circ$         & 77.16 & 77.97 & 79.52 & \underline{79.79} & \textbf{81.16} & 66.02 & 65.92 & \underline{67.65} & 66.39 & \textbf{71.67} & 37.30 & 38.03 & 39.07          & \underline{39.74}          & \textbf{45.17} \\
$\bullet$        & $\circ$         & $\circ$          & $\circ$         & 87.32 & 87.40 & 86.92 & \underline{88.07} & \textbf{89.83} & 69.19 & 68.79 & 69.92 & \underline{72.45} & \textbf{75.01} & 38.15 & 42.37 & \underline{42.82}          & 42.29          & \textbf{47.75} \\
$\circ$        & $\circ$         & $\bullet$          & $\bullet$         & 87.74 & 87.81 & 88.35 & \underline{88.70} & \textbf{89.37} & 83.45 & 84.75 & \underline{86.29} & 84.55 & \textbf{87.06} & 75.93 & 79.39 & 80.13  & \underline{80.49}          & \textbf{83.46} \\
$\circ$        & $\bullet$         & $\bullet$          & $\circ$         & 81.12 & 81.75 & \underline{83.10} & 83.09 & \textbf{84.05} & 83.40 & 83.56 & \underline{85.83} & 83.18 & \textbf{86.56} & 78.01 & 80.07 & \underline{81.72}          & 80.82          & \textbf{82.24} \\
$\bullet$        & $\bullet$         & $\circ$          & $\circ$         & 89.73 & 89.79 & 89.80 & \underline{90.10} & \textbf{91.12} & 73.07 & 73.84 & \underline{74.44} & 73.72 & \textbf{77.51} & 40.98 & 45.92 & 46.76          & \underline{49.50}          & \textbf{56.62} \\
$\circ$        & $\bullet$         & $\circ$          & $\bullet$         & 87.73 & 87.81 & 87.93 & \underline{88.72} & \textbf{89.11} & 73.13 & 73.41 & 72.88 & \underline{73.43} & \textbf{75.58} & 45.65 & 46.76 & 47.28          & \underline{53.05}          & \textbf{57.48} \\
$\bullet$        & $\circ$         & $\circ$          & $\bullet$         & 89.87 & 89.85 & 90.05 & \underline{90.38} & \textbf{91.09} & 74.14 & 74.61 & 74.48 & \underline{75.66} & \textbf{77.05} & 49.32 & 48.64 & 49.48          & \underline{54.36}          & \textbf{58.18} \\
$\bullet$        & $\circ$         & $\bullet$          & $\circ$         & 89.89 & 89.34 & 89.95 & \underline{90.64} & \textbf{91.34} & 84.65 & 84.78 & \underline{86.59} & 85.96 & \textbf{87.19} & 76.67 & \underline{81.89} & 81.22          & 80.99          & \textbf{83.95} \\
$\bullet$        & $\bullet$         & $\bullet$          & $\circ$         & 90.69 & 90.11 & 90.61 & \underline{91.08} & \textbf{91.58} & 85.07 & 85.22 & \underline{86.74} & 85.75 & \textbf{87.16} & 76.81 & 82.09 & 81.83          & \underline{82.27}          & \textbf{84.08} \\
$\bullet$        & $\bullet$         & $\circ$          & $\bullet$         & 90.60 & 90.57 & 90.63 & \underline{91.05} & \textbf{91.30} & 75.19 & 75.58 & \underline{75.81} & 75.69 & \textbf{77.80} & 49.92 & 50.29 & 51.05          & \underline{53.87}          & \textbf{59.88} \\
$\bullet$        & $\circ$         & $\bullet$          & $\bullet$         & 90.68 & 90.38 & 90.84 & \underline{91.33} & \textbf{91.68} & 84.97 & 85.32 & \underline{86.41} & 86.04 & \textbf{87.18} & 77.12 & 78.73 & 80.01          & \underline{81.14}          & \textbf{84.11} \\
$\circ$        & $\bullet$         & $\bullet$          & $\bullet$         & 88.25 & 88.16 & 88.87 & \underline{89.01} & \textbf{89.72} & 83.47 & 84.17 & \underline{86.47} & 84.34 & \textbf{87.30} & 76.99 & 79.29 & \underline{82.06}          & 81.24          & \textbf{84.24} \\
$\bullet$        & $\bullet$         & $\bullet$          & $\bullet$         & 91.11 & 90.64 & 91.03 & \underline{91.60} & \textbf{91.79} & 85.21 & 84.61 & \underline{86.40} & 85.75 & \textbf{87.03} & 78.00 & 79.92 & 80.98          & \underline{83.00} & \textbf{83.23} \\ \hline
\multicolumn{4}{c|}{Average}                   & 86.98 & 87.08 & 87.58 & \underline{88.09} & \textbf{88.90} & 78.23 & 78.69 & \underline{79.67} & 79.24 & \textbf{81.58} & 61.47 & 64.08 & 64.87          & \underline{66.42}          & \textbf{69.71} \\ \hline

\end{tabular}
}
\end{table*}

\begin{table*}
\caption{DSC(\%) comparison with RFNet~\cite{RFNet_fusion4}, mmFormer (mmF)~\cite{mmFormer_fusion6}, MAVP~\cite{MAVP_fusion7} and GSS~\cite{GSS_union2} on BraTS2018. Available modalities are denoted by $\bullet$, and missing modalities are represented by $\circ$. The best DSC results are indicated in bold, and the second-best DSC results are underlined.}
\label{tab3_result_comparison2018}
\centering
\resizebox{\linewidth}{!}{
\begin{tabular}{cccc|ccccc|ccccc|ccccc}
\hline
\multicolumn{4}{c|}{Modality}  & \multicolumn{5}{c|}{WT}  & \multicolumn{5}{c|}{TC}  & \multicolumn{5}{c}{ET}  \\ \hline
$fl$        & $t1$        & $tc$        & $t2$        & RFNet & mmF   & MAVP  & GSS   & Ours           & RFNet & mmF   & MAVP  & GSS   & Ours           & RFNet & mmF   & MAVP           & GSS            & Ours           \\ \hline
$\circ$        & $\circ$         & $\circ$          & $\bullet$         & 84.30 & 85.64 & 85.88 & \underline{86.40} & \textbf{87.22} & 67.62 & 68.04 & 68.04 & \underline{69.43} & \textbf{72.12} & 40.17 & 40.32 & 40.93 & \textbf{45.76} & \underline{45.07} \\
$\circ$        & $\circ$         & $\bullet$          & $\circ$         & 74.93 & 77.11 & 77.40 & \underline{78.47} & \textbf{81.99} & 80.99 & 80.68 & 80.49 & \underline{82.32} & \textbf{84.81} & 69.43 & 76.02 & 75.87 & \underline{77.10} & \textbf{78.98} \\
$\circ$        & $\bullet$         & $\circ$          & $\circ$         & 74.68 & 77.01 & 77.72 & \underline{78.79} & \textbf{80.57} & 64.42 & 64.61 & 65.28 & \underline{67.47} & \textbf{70.49} & 34.43 & 32.74 & 34.24 & \underline{42.37} & \textbf{45.63} \\
$\bullet$        & $\circ$         & $\circ$          & $\circ$         & 86.46 & 87.17 & 87.29 & \underline{87.65} & \textbf{89.53} & 64.89 & 64.53 & 64.75 & \underline{68.60} & \textbf{71.99} & 33.92 & 35.62 & 35.22 & \textbf{42.88} & \underline{42.81} \\
$\circ$        & $\circ$         & $\bullet$          & $\bullet$         & 86.39 & 86.87 & 87.04 & \underline{87.94} & \textbf{88.14} & 83.27 & 82.06 & 82.15 & \underline{84.35} & \textbf{84.92} & 73.01 & 75.82 & 77.73 & \textbf{79.39} & \underline{77.77}\\
$\circ$        & $\bullet$         & $\bullet$          & $\circ$         & 78.59 & 80.52 & 81.03 & \underline{81.90} & \textbf{83.51} & 82.22 & 82.03 & 82.29 & \underline{83.71} & \textbf{85.00} & 70.73 & 75.63 & 76.47 & \underline{77.90} & \textbf{80.48} \\
$\bullet$        & $\bullet$         & $\circ$          & $\circ$         & 88.78 & 88.85 & 89.03 & \underline{89.56} & \textbf{90.29} & 71.59 & 70.77 & 72.77 & \underline{73.81} & \textbf{75.68} & 39.68 & 39.36 & 39.70 & \underline{47.34} & \textbf{51.08} \\
$\circ$        & $\bullet$         & $\circ$          & $\bullet$         & 86.15 & 86.92 & 86.94 & \underline{87.48} & \textbf{87.87} & 70.89 & 70.42 & 70.58 & \underline{73.24} & \textbf{75.28} & 41.42 & 42.45 & 44.37 & \underline{48.98} & \textbf{49.72} \\
$\bullet$        & $\circ$         & $\circ$          & $\bullet$         & 89.12 & 89.48 & 89.59 & \underline{89.93} & \textbf{90.50} & 70.82 & 70.44 & 70.47 & \underline{73.38} & \textbf{74.80} & 43.77 & 43.46 & 44.51 & \underline{48.59} & \textbf{51.08} \\
$\bullet$        & $\circ$         & $\bullet$          & $\circ$         & 89.17 & 89.51 & 89.68 & \underline{89.90} & \textbf{90.11} & 82.94 & 81.98 & 82.34 & \underline{83.71} & \textbf{85.00} & 72.84 & 76.33 & 76.93 & \underline{77.84} & \textbf{78.79} \\
$\bullet$        & $\bullet$         & $\bullet$          & $\circ$         & 89.71 & 89.66 & 89.74 & \textbf{90.25} & \underline{90.19} & 83.77 & 82.79 & 83.06 & \underline{84.73} & \textbf{85.27} & 73.17 & 76.69 & 77.64 & \underline{78.42} & \textbf{80.08} \\
$\bullet$        & $\bullet$         & $\circ$          & $\bullet$         & 89.68 & 89.76 & 89.81 & \underline{90.23} & \textbf{90.25} & 73.09 & 72.01 & 72.20 & \underline{75.37} & \textbf{76.21} & 44.79 & 44.00 & 44.06 & \underline{50.17} & \textbf{53.12} \\
$\bullet$        & $\circ$         & $\bullet$          & $\bullet$         & 90.06 & 90.11 & 90.22 & \textbf{90.73} & \underline{90.48} & 83.54 & 82.05 & 82.12 & \underline{84.42} & \textbf{84.93} & 73.13 & 75.84 & 77.50 & \underline{78.69} & \textbf{79.38} \\
$\circ$        & $\bullet$         & $\bullet$          & $\bullet$         & 86.78 & 87.07 & 87.18 & \underline{88.04} & \textbf{88.30} & 83.97 & 82.45 & 82.43 & \underline{84.56} & \textbf{85.10} & 72.56 & 76.18 & 77.54 & \underline{78.51} & \textbf{79.47} \\
$\bullet$        & $\bullet$         & $\bullet$          & $\bullet$         & 90.26 & 90.14 & 90.18 & \textbf{90.74} & \underline{90.49} & 84.02 & 82.36 & 82.41 & \underline{84.61} & \textbf{85.29} & 73.21 & 76.43 & 77.24 & \underline{78.33} & \textbf{80.01} \\ \hline
\multicolumn{4}{c|}{Average}                   & 85.67 & 86.38 & 86.60 & \underline{87.20} & \textbf{87.96} & 76.53 & 75.82 & 76.00 & \underline{78.25} & \textbf{79.79} & 57.12 & 59.12 & 60.01 & \underline{63.49} & \textbf{64.90}\\ \hline

\end{tabular}
}
\end{table*}

\begin{table*}
\caption{DSC(\%) comparison with RFNet~\cite{RFNet_fusion4}, mmFormer (mmF)~\cite{mmFormer_fusion6}, MAVP~\cite{MAVP_fusion7} and GSS~\cite{GSS_union2} on BraTS2015. Available modalities are denoted by $\bullet$, and missing modalities are represented by $\circ$. The best DSC results are indicated in bold, and the second-best DSC results are underlined.}
\label{tab3_result_comparison2015}
\centering
\resizebox{\linewidth}{!}{
\begin{tabular}{cccc|ccccc|ccccc|ccccc}
\hline
\multicolumn{4}{c|}{Modality}  & \multicolumn{5}{c|}{WT}  & \multicolumn{5}{c|}{TC}  & \multicolumn{5}{c}{ET}  \\ \hline
$fl$        & $t1$        & $tc$        & $t2$        & RFNet & mmF   & MAVP  & GSS   & Ours           & RFNet & mmF   & MAVP  & GSS   & Ours           & RFNet & mmF   & MAVP           & GSS            & Ours           \\ \hline
$\circ$        & $\circ$         & $\circ$          & $\bullet$         & 86.89 & 86.99 & 88.51 & \underline{88.55} & \textbf{89.54} & 63.81 & 60.07 & 63.11 & \underline{65.77} & \textbf{67.80} & 40.07 & 38.55 & 38.77 & \underline{41.48} & \textbf{47.94}\\
$\circ$        & $\circ$         & $\bullet$          & $\circ$         & 74.95 & 74.22 & 75.94 & \underline{77.90} & \textbf{78.79} & 72.64 & 72.37 & \underline{73.54} & 73.06 & \textbf{77.19} & \textbf{81.40} & 76.48 & 76.96 & \underline{81.00} & 75.90 \\
$\circ$        & $\bullet$         & $\circ$          & $\circ$         & 74.20 & 69.52 & 73.10 & \underline{77.02} &\textbf{79.15} & 61.27 & 56.88 & 61.16 & \underline{66.30} & \textbf{66.71} & 29.44 & 33.26 & 34.58 & \textbf{42.67} & \underline{39.30} \\
$\bullet$        & $\circ$         & $\circ$          & $\circ$         & 86.91 & 87.28 & 87.14 & \underline{88.04} & \textbf{91.55} & 58.71 & 52.68 & 57.20 & \underline{63.60} & \textbf{66.98} & 35.23 & 35.00 & 36.59 & \underline{38.86} & \textbf{48.01} \\
$\circ$        & $\circ$         & $\bullet$          & $\bullet$       & 88.39 & 88.69 & 88.38 & \underline{89.39} & \textbf{90.45} & 77.50 & 77.03 & 78.09 & \underline{79.78} & \textbf{81.99} & \underline{86.97} & 81.85 & 79.31 & \textbf{86.98} & 81.70 \\
$\circ$        & $\bullet$         & $\bullet$          & $\circ$       & 78.13 & 74.70 & 77.18 & \textbf{80.08} & \underline{80.02} & 74.06 & 71.04 & 73.02 & \underline{74.29} & \textbf{77.73} & \textbf{82.48} & 76.25 & 77.15 & \underline{82.05} & 81.40 \\
$\bullet$        & $\bullet$         & $\circ$          & $\circ$         & 88.51 & 87.55 & 88.90 & \underline{89.72} & \textbf{91.62} & 66.88 & 62.35 & 67.26 & \underline{70.47} & \textbf{70.48} & 40.95 & 41.32 & 42.04 & \underline{45.42} & \textbf{46.67} \\
$\circ$        & $\bullet$         & $\circ$          & $\bullet$         & 88.25 & 87.97 & \underline{89.68} & 89.36 & \textbf{90.42} & 67.24 & 67.44 & 67.88 & \underline{69.53} & \textbf{71.51} & 40.58 & 39.03 & 40.95 & \underline{46.69} & \textbf{48.11} \\
$\bullet$        & $\circ$         & $\circ$          & $\bullet$         & 89.62 & 89.88 & 89.78 & \underline{90.82} & \textbf{91.85} & 68.74 & 65.07 & 67.02 & \textbf{70.64} & \underline{69.82} & 44.64 & 42.12 & 40.22 & \underline{49.03} & \textbf{51.05} \\
$\bullet$        & $\circ$         & $\bullet$          & $\circ$         & 88.45 & 89.83 & 88.79 & \underline{90.14} & \textbf{91.53} & 79.30 & 76.48 & 79.07 & \underline{81.74} & \textbf{82.74} & \underline{86.15} & 84.46 & 85.25 & \textbf{86.84} & 84.52 \\
$\bullet$        & $\bullet$         & $\bullet$          & $\circ$         & 88.75 & 89.31 & 89.31 & \underline{90.28} & \textbf{91.72} & 80.46 & 77.86 & 79.75 & \underline{81.59} & \textbf{84.14} & \underline{87.30} & 83.25 & 83.20 & \textbf{87.62} & 85.66 \\
$\bullet$        & $\bullet$         & $\circ$          & $\bullet$         & 89.93 & 89.64 & 90.50 & \underline{90.83} & \textbf{92.13} & 69.75 & 68.55 & 68.76 & \textbf{72.07} & \underline{70.94} & 44.21 & 44.64 & 44.47 & \textbf{52.05} & \underline{47.70} \\
$\bullet$        & $\circ$         & $\bullet$          & $\bullet$         & 90.07 & 90.97 & \underline{91.30} & 91.16 & \textbf{92.43} & 79.29 & 79.78 & 80.75 & \underline{81.88} & \textbf{82.97} & \underline{87.34} & 86.05 & 86.24 & \textbf{87.64} & 84.81 \\ 
$\circ$        & $\bullet$         & $\bullet$          & $\bullet$         & 88.41 & 88.08 & 88.54 & \underline{89.45} & \textbf{90.78} & 79.18 & 80.38 & \underline{80.93} & 79.99 & \textbf{83.30} & \underline{87.47} & 79.03 & 79.35 & \textbf{87.50} & 84.36 \\
$\bullet$        & $\bullet$         & $\bullet$          & $\bullet$         & 90.49 & 90.41 & \underline{91.32} & 91.11 & \textbf{92.41} & 80.16 & 81.42 & 81.53 & \underline{81.72} & \textbf{83.94} & \underline{87.68} & 85.24 & 85.82 & \textbf{87.97} & 85.39 \\ \hline
\multicolumn{4}{c|}{Average}                   & 86.13 & 85.67 & 86.56 & \underline{87.59}  & \textbf{88.96} & 71.93 & 69.96 & 71.94 & \underline{74.16} & \textbf{75.88} & 64.13 & 61.77 & 62.06 & \textbf{66.92} & \underline{66.17} \\ \hline

\end{tabular}
}
\end{table*}

\begin{figure}
\centerline{\includegraphics[width=\columnwidth]{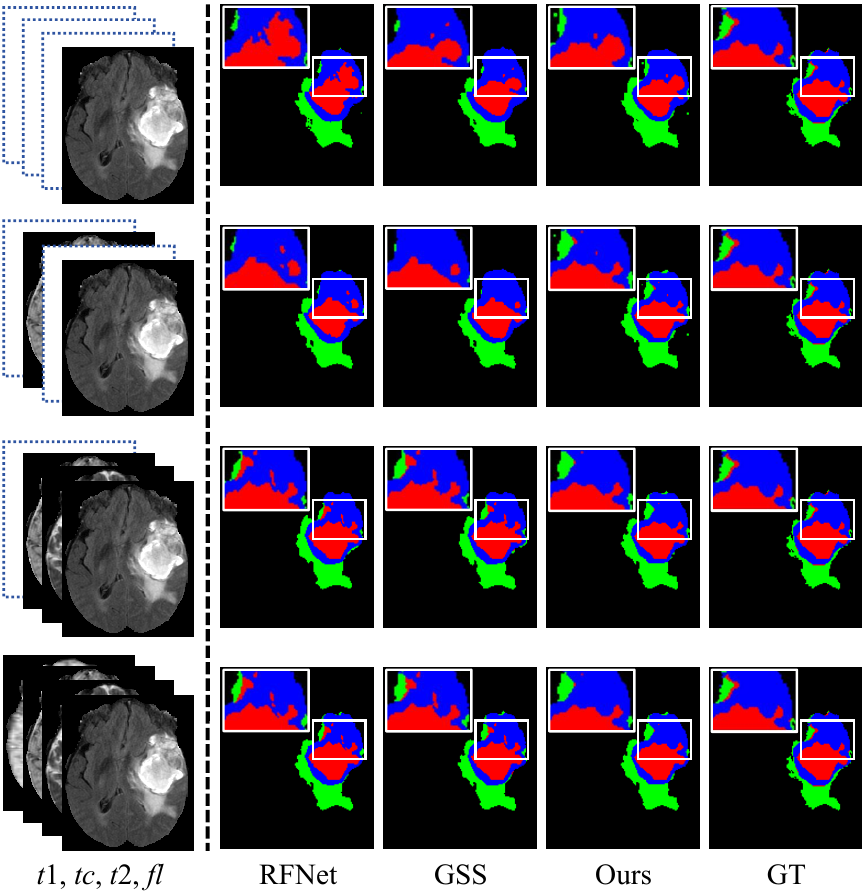}}
\caption{Visualization comparisons with SOTA methods on different missing modality scenarios. Green, Blue and Red represent WT, ET and TC, respectively. Left: four modality input images when missing scenarios. Right: predictions of RFNet, GSS and DeMoSeg, and the corresponding ground truth.} 
\label{fig3_visualization}
\end{figure}

\begin{figure*}
\centering
\includegraphics[width=\textwidth]{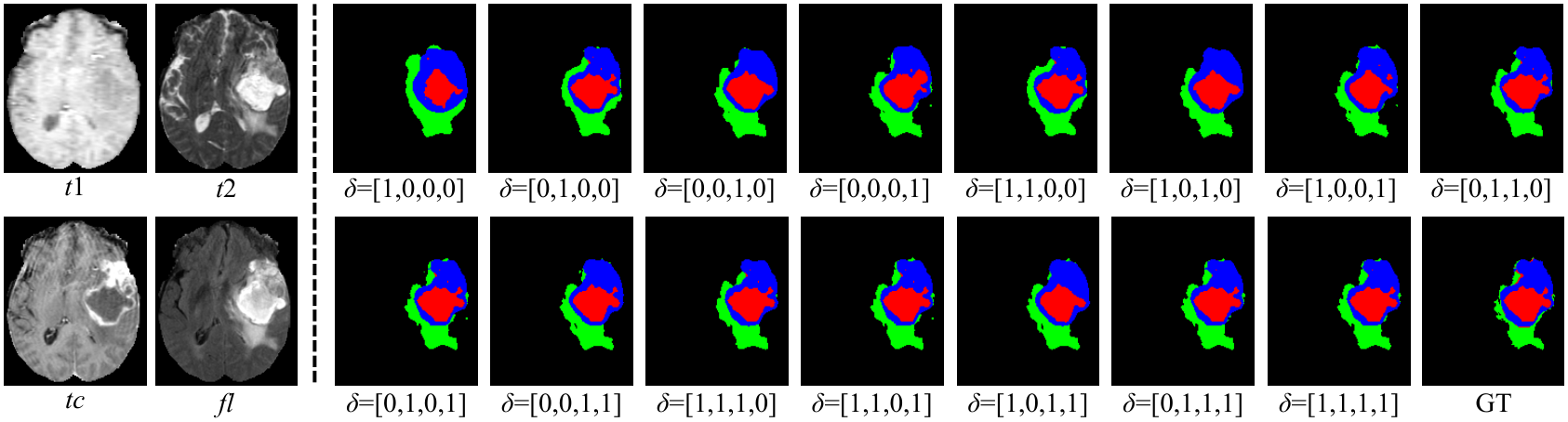}
\caption{The segmentation results of DeMoSeg on different missing modality scenarios. Green, Blue and Red represent WT, ET and TC, respectively. Left: input images of four modalities. Right: the segmentation results when different missing modality scenario and corresponding ground truth. The $\delta = [\delta_{t1}, \delta_{tc}, \delta_{t2}, \delta_{fl}]$ indicates the present and absent input modalities.} 
\label{fig4_visualization}
\end{figure*}

\textbf{Visualization.} We also visualize some cases in Fig.~\ref{fig3_visualization} and Fig.~\ref{fig4_visualization}. Fig.~\ref{fig3_visualization} shows the comparison of segmentation results between DeMoSeg and other SOTA methods under different missing modalities. The white boxes in the figure highlight the differences between prediction and the ground truth, which indicates that our DeMoSeg produces closer segmentation to the ground truth and demonstrates more robust performance under different missing modality conditions.
Fig.~\ref{fig4_visualization} shows the prediction of DeMoSeg under various scenarios, which shows that DeMoSeg can achieve decent segmentation when full modality and missing modality situations, even with only one available modality.

\textbf{Efficiency Comparison.} 
To better illustrate the effectiveness of different methods, we have compiled statistics on the model parameters, floating point operations (FLOPs) and GPU memory consumption of the \textbf{Enabling Module} proposed by comparison methods.
For fair comparison, a batch size of $1$ is utilized for all methods. Based on the default training setting, we set the patch size for RFNet to $80\times 80\times 80$ and $128\times 128\times 128$ for mmF, MAVP and DeMoSeg. We define an efficiency factor $P$ to better quantify the effectiveness of different methods, which can be formulated as:
\begin{equation}
    P = \frac{\Delta DSC (\%)}{\lambda \cdot Param(M) + \mu \cdot FLOPs(G)/\eta^3}
\end{equation}
where $Param$, $FLOPs$ are calculated on \textbf{Enabling Module}, $\eta$ is the scaling factor for patch size, $80$ corresponds to $1$ and $128$ corresponds to $1.6$, $\lambda$ and $\mu$ are adjustable weight factors for $Param$ and $FLOPs$, respectively, with $\lambda =\mu = 0.5$ here. The efficiency factor $P$ reflects the impact of $Param$ and $FLOPs$ on the improvement in DSC. A higher value of $P$ indicates better performance of enabling module, which means that the proposed module achieves significant improvement with fewer $Param$ and $FLOPs$.
From the results in Table~\ref{tab_efficiency}, it demonstrates that our constructed modules, \textit{i.e.}, Feature Decoupling, CSSA and RCR, are light-weighted and effective, with lower parameters and higher effectiveness. 


\begin{table}[htbp]
\caption{Efficiency comparison with SOTA methods. All results are calculated using a batch size of 1, with a patch size of $80\times 80\times 80$ for RFNet and $128\times 128\times 128$ for the others. Param, FLOPs and GPU represent the parameters, FLOPs and GPU consumption of the \textbf{enabling module}, respectively. $\Delta$DSC is the improvement of average DSC compared with corresponding baseline on BraTS2020.}
\label{tab_efficiency}
    \centering
    \begin{tabular}{c|ccccc} \hline
    Methods & Param $\downarrow$ &  FLOPs $\downarrow$ & GPU $\downarrow$ & $\Delta$DSC $\uparrow$ & $P \uparrow$ \\ \hline
    RFNet & 6.9M & 162G & 7.3GB & 6.01\% & 0.071 \\
    mmF & 27M & 58G & 11GB & 0.72\% & 0.035 \\
    MAVP & 7.9M & - &  5.4GB & 0.76\% & - \\
    DeMoSeg & 0.3M & 176G & 1.6GB & 4.10\% & 0.189 \\
    \hline
    \end{tabular}
\end{table}

\subsection{Ablation Study.}
In this section, we firstly explore the roles of our proposed modules in DeMoSeg. Then, we also investigate the impact of the feature compensation orders in RCR on tumor segmentation results. Finally, we examine the way and location of constraints between Self- and Mutual-features during training.

\textbf{Effectiveness of key components.} We firstly investigate the effectiveness of three key components, \textit{i.e.}, Feature Decoupling (FD), Channel-wised Sparse Self-Attention (CSSA) and Radiologist-mimic Cross-modality expression Relationships (RCR), by training eight variants of DeMoSeg using the components with different combinations, and the comparison results of the variants on BraTS2020 are listed in Table~\ref{tab4_ablation_study}. 
Note that, we utilize a convolution to replace the RCR and employ the whole entangled feature as the Mutual-features for the ablation study of only using FD and RCR, respectively.
Results from the first four rows of Table~\ref{tab4_ablation_study} indicate a notable improvement of the three components compared to the baseline, demonstrating the effectiveness of these modules on incomplete multi-modal brain tumor segmentation.
The complete DeMoSeg outperforms the variant where FD is absent (the fifth vs. last row), which implies that the learning burden for feature adaptation is alleviated. CSSA creates a sparse communication path between Self- and Mutual-features. With such path, the features turn out to learn better to yield higher accuracy. Therefore, CSSA can bring mutual guidance across the two type of features (the sixth vs. last row). Furthermore, clinical knowledge-based feature compensation yields more robust fused features especially for better distinguishing the hard region ET (see the last two rows). 

\begin{table}[htbp]
\caption{Ablation study of the proposed components of DeMoSeg on BraTS2020.}
\setlength{\tabcolsep}{8pt}
\label{tab4_ablation_study}
    \centering
    \begin{tabular}{ccc|ccc} \hline
    \multirow{2}{*}{FD} & \multirow{2}{*}{CSSA} & \multirow{2}{*}{RCR} & \multicolumn{3}{c}{DSC(\%)}           \\ \cline{4-6}
    & & &  WT &  TC &  ET \cr  \hline
    \ding{55} & \ding{55} & \ding{55} &  86.60 &  78.65 & 62.64 \cr
    \checkmark & \ding{55} & \ding{55} & 88.24 & 80.61 & 66.74 \cr
    \ding{55} & \checkmark & \ding{55} & 88.18 & 79.68 & 67.32 \cr
    \ding{55} & \ding{55} & \checkmark & 88.04 & 80.39 & 66.60 \cr
    \ding{55} & \checkmark & \checkmark & 88.78 & 80.72 & 68.32 \cr
    \checkmark & \ding{55} & \checkmark & 88.80 & 81.42 & 68.73 \cr
    \checkmark & \checkmark & \ding{55} & 88.83 & 81.47 & 67.63 \cr
     \checkmark &  \checkmark &  \checkmark &  \textbf{88.90} &  \textbf{81.58} &  \textbf{69.71} \cr 
    \hline
    \end{tabular}
\end{table}

\textbf{Features compensation orders in RCR.} In RCR, we have constructed three types of feature compensation relationships incorporating the clinical knowledge, the performance of single-modality segmentation and other prior information, \textit{i.e.}, 
\begin{itemize}
    \item \makebox[1.5cm][l]{\textit{primary}:}   $\{t1\leftrightarrow tc$, $t2\leftrightarrow fl\}$ $\triangleq \mathcal{I}$
    \item \makebox[1.5cm][l]{\textit{secondary}:} $\{t1\leftrightarrow t2$, $tc\leftrightarrow fl\}$ $\triangleq \mathcal{II}$
    \item \makebox[1.5cm][l]{\textit{tertiary}:}  $\{t1\leftrightarrow fl$, $tc\leftrightarrow t2\}$ $\triangleq \mathcal{III}$
\end{itemize}

To explore the rationale of the three compensation relationships, we conduct six variants of DeMoSeg by altering the importance order of $\mathcal{I}$, $\mathcal{II}$ and $\mathcal{III}$ in RCR. The results of RCR order on BraTS2020 are shown in Table~\ref{tab_ablation_study_RCR}, which demonstrates that the feature compensation order $\mathcal{I} \rightarrow \mathcal{II} \rightarrow \mathcal{III}$ is optimal for the missing modalities brain tumor segmentation. This ablation study also proves the rationale and effectiveness of RCR in constructing the pseudo full modality features $\mathbf{F}^{fused}$ without introducing any computational burden.

\begin{table}[htbp]
\caption{Ablation study of the features compensation orders in RCR on BraTS2020. $A \rightarrow B \rightarrow C$ represents the current feature compensation order is as follow: A, B and C are primary, secondary and tertiary relationship, respectively.}
\setlength{\tabcolsep}{8pt}
\label{tab_ablation_study_RCR}
    \centering
    \begin{tabular}{c|ccc} \hline
    \multirow{2}{*}{Compensation Order} & \multicolumn{3}{c}{DSC(\%)}           \\ \cline{2-4}
    &  WT &  TC &  ET \cr  \hline
    $\mathcal{III} \rightarrow \mathcal{II} \rightarrow \mathcal{I}$ & 88.81 & 81.26 & 68.74 \\
    $\mathcal{III} \rightarrow \mathcal{I} \rightarrow \mathcal{II}$ & 88.87 & 81.57 & 68.44 \\
    $\mathcal{II} \rightarrow \mathcal{III} \rightarrow \mathcal{I}$ & 88.65 & 81.25 & 67.87 \\
    $\mathcal{II} \rightarrow \mathcal{I} \rightarrow \mathcal{III}$ & 88.64 & \textbf{81.65} & 68.29 \\
    $\mathcal{I} \rightarrow \mathcal{III} \rightarrow \mathcal{II}$ & 88.77 & 81.59 & 69.56 \\
    $\mathcal{I} \rightarrow \mathcal{II} \rightarrow \mathcal{III}$  &  \textbf{88.90} &  81.58 &  \textbf{69.71} \\
    \hline
    \end{tabular}
\end{table}

\textbf{Feature alignment constraint in FD.} In the process of feature decoupling, we also conducted experiments to explore whether to use and the location of alignment constraint $\mathcal{L}_{kd}$ between Self- and Mutual-expression. 

(1) Whether to use constraint: Originally, we utilize the knowledge distillation in DeMoSeg to align the features between $u'_{(\mathcal{M}\backslash\{m\})\rightarrow m}$ and $s'_m$. We also want to verify the effect of this constraint.

(2) The location of constraint: We want to clarify whether we should add the feature constraint before or after CSSA, that is, whether the constrained features should be ($u'_{(\mathcal{M}\backslash\{m\})\rightarrow m}, s'_m$) or ($u_{(\mathcal{M}\backslash\{m\})\rightarrow m}, s_m$).

\begin{table}[htbp]
\caption{Ablation study of the constraint in FD on BraTS2020.}
\setlength{\tabcolsep}{8pt}
\label{tab_ablation_study_FD}
    \centering
    \begin{tabular}{c|ccc} \hline
    \multirow{2}{*}{Constraint} & \multicolumn{3}{c}{DSC(\%)}           \\ \cline{2-4}
     &  WT &  TC &  ET \cr  \hline
    w/o $\mathcal{L}_{kd}$ & 88.87 & 81.34 & 68.52 \cr \hline 
    w/ $\mathcal{L}_{kd}$, after CSSA & 88.65 & 81.39 & 68.72 \cr
    w/ $\mathcal{L}_{kd}$, before CSSA & \textbf{88.90} &  \textbf{81.58} &  \textbf{69.71} \cr 
    \hline
    \end{tabular}
\end{table}

From Table~\ref{tab_ablation_study_FD}, the results of \textbf{KL} constraint before CSSA are superior than other constraint variants. 
We can draw the following conclusions: Firstly, using constraints in FD allows the Mutual-features to perform better during feature compensation. Secondly, applying constraint before CSSA yields better results, and we guess post-constraint conflicts with CSSA. 

\section{Conclusion}
In this work, we propose a novel efficient and lightweight framework DeMoSeg for incomplete multi-modal brain tumor segmentation. Instead of learning highly-entangled fused features, we decouple the Self-feature and Mutual-features of each modality for ego and other modalities to reduce the learning burden. Moreover, we prove that the proposed CSSA can enhance mutual-guidance between the two kinds of features as well as prevent them from being highly entangled again. Besides, the clinical-knowledge based feature compensation strategy RCR effectively constructs pseudo full-modality features from available modalities under various scenarios. Extensive experimental results on BraTS2020, BraTS2018 and BraTS2015 have demonstrated the superiority and reliability of our DeMoSeg over previous state-of-the-art methods, especially for quite challenging scenarios where more than one modalities are unavailable. In the future, we intend to augment feature interactions to provide more robust fused features, and introduce some feature style transformation mechanism to better guide the knowledge distillation between Self- and Mutual-features. 



\bibliographystyle{IEEEtran}
\bibliography{IEEEabrv,ref}

\end{document}